\documentclass{article}

\usepackage{microtype}
\usepackage{graphicx}
\usepackage{subfigure}
\usepackage{booktabs} 
\usepackage{array}

\usepackage{hyperref}  

\usepackage[accepted]{icml2025}

\usepackage{amsmath}
\usepackage{amssymb}
\usepackage{mathtools}
\usepackage{amsthm}

\usepackage[capitalize,noabbrev]{cleveref}

\usepackage{multirow}
\theoremstyle{plain}

\theoremstyle{definition}

\theoremstyle{remark}

\usepackage[textsize=tiny]{todonotes}

\icmltitlerunning{}

\begin{document}

\twocolumn[
\icmltitle{EDGE: Engine for Deterministic Graph Evaluation through Conversation Simulation from Graph Structured DSL Configuration}



\icmlsetsymbol{equal}{*}

\begin{icmlauthorlist}
\icmlauthor{Ram Kulathumani}{equal,comp}
\icmlauthor{Pushkar Nagar}{equal,comp}
\icmlauthor{Regunathan Radhakrishnan}{equal,comp}
\icmlauthor{Anupam Tripathi}{equal,comp}
\icmlauthor{Xiangbo Mao}{equal,comp}
\icmlauthor{Roshanak Omrani}{equal,comp}
\icmlauthor{Keshav Somani}{equal,comp}
\icmlauthor{Shwet Kamal Mishra}{equal,comp}
\icmlauthor{Shayna Lurya}{equal,comp}
\icmlaffiliation{comp}{Salesforce, Palo Alto, USA}
\icmlcorrespondingauthor{Ram Kulathumani}{rkulathumani@salesforce.com}
\end{icmlauthorlist}
\icmlkeywords{Machine Learning, ICML}

\vskip 0.3in
]



\printAffiliationsAndNotice{Authors are listed alphabetically. \icmlEqualContribution} 

\begin{abstract}
As agentic systems evolve into complex multi agent orchestration workflows, there is a growing and critical need for systematic frameworks that measures an agent's behavioral consistency and determinism. In this paper, we introduce a formal evaluation methodology that is grounded in AgentGraph, a planner powered by a domain specific language that represents agent reasoning through a dynamically adjustable directed graph. We leverage this structural formalism and utilize graph traversal algorithms that exhaustively enumerate conversational paths, forming a comprehensive evaluation set that captures the agent's complete behavioral space. We then systematically replay these reproducible trajectories to compare observed outputs and state transitions against the intended DSL specification. To quantify reliability, we define novel metrics that measure response and trajectory determinism, structural adherence and semantic consistency across both exact replays and their linguistic variants. Our system's results demonstrate that agents configured using frameworks like AgentGraph and LangGraph with explicitly structured node transitions show superior determinism over agents that are not configured with controlled transitions.
\end{abstract}

\section{Introduction}
\label{submission}
The growing complexity of agentic systems has amplified the need for systematic evaluation frameworks that can measure determinism, reproducibility and behavioral consistency across conversations. Agentic architectures have evolved from monolithic systems to multi step, multi agent orchestration workflows. Consistent reasoning and decision making in these systems have become both critical and non trivial. In this paper, we introduce a novel graph exhaustive evaluation framework for measuring behavioral determinism in agentic systems. The evaluations are grounded in the structural representation of agents specified using a domain specific language configuration known as the \textbf{AgentGraph}, introduced in \textit{Agentforce}. AgentGraph formalizes an agent’s reasoning and conversational topology as a dynamically adjustable directed graph where edges could be controlled using state variables. This allows users to design predictable agent interactions. We leverage this formalism to algorithmically enumerate all possible conversation flows using a Depth First Traversal of the AgentGraph. This forms our comprehensive evaluation dataset that captures the full behavior space of the Agent under evaluation. Each graph traversal represents a distinct yet reproducible conversational path that could be independently executed and evaluated. We systematically compare the agent’s observed outputs and state transitions as specified in the DSL graph across multiple runs for each graph traversal. We then quantify determinism metrics that measure response stability, structural adherence and semantic consistency. This approach provides an interpretable and extensible framework for diagnosing stochasticity and emergent variability in complex agent systems. Our contributions are threefold:
\begin{itemize}
    \item we introduce a formal methodology to operationalize determinism evaluation for agentic systems that use graph based DSL configurations
    \item we present an automated evaluation set generation pipeline that enumerates conversational path for large scale and reproducible evaluation
    \item we define novel determinism metrics that capture both lexical and structural consistency in agent behavior.
\end{itemize}
Together, these contributions establish a framework for determinism evaluation of graph based Agentic Systems  with complex reasoning expressed on authoring platforms like AgentForce, LangGraph etc. To evaluate agent determinism, we simulate multi turn conversations to exhaustively test different graph paths. The resulting metrics reveal how consistently agents maintain trajectories when faced with either exact replays or varied user inputs for a single task. Specifically, for an agent configuration intentionally designed with care for node transitions, we were able to show better determinism against another agent that did not have careful consideration on node transitions.

\section{Related Work}
The rapid evolution of autonomous agents, with increasingly complex logic and growing number of tools, has necessitated a shift from static, single-turn benchmarks to complex trajectory-based assessment. In addition to final response and step-wise evaluation, some platforms provide pipelines and metrics that support benchmarking the sequence of agent's decisions with respect to an expected optimal path, as well as assessing the sequence in the absence of such reference path \cite{yehudai2025}. We situate our work within three primary research domains: Empirical Observability, Stochastic User Simulation, and Structured Agent Orchestration.
\subsection{Empirical Observability and Trajectory Evaluation}
Current state-of-the-practice frameworks, most notably LangSmith \yrcite{langsmith} and Arize Phoenix \yrcite{arizephoenix}, focus on post-hoc trace analysis. These systems utilize distributed tracing to capture nested execution logs, or ``trajectories,'' which are then evaluated using LLM-as-a-Judge heuristics \cite{khatchadourian2025}. While vital for production monitoring, these approaches are fundamentally reactive and probabilistic. They evaluate only the specific paths triggered by real-world user traffic or fixed datasets, leaving the vast majority of an agent's logic space unprobed. Our work addresses this ``Cold-Start Path Problem'' by shifting from an empirical analysis of observed traces to a formal verification of the underlying graph topology.

\subsection{Stochastic User Simulation and Benchmarking}
To move beyond static input-output pairs, frameworks like OpenEvals \yrcite{openevals} and AgentEvals \yrcite{agentevals} employ closed-loop simulations where a ``challenger agent'' interacts with the system under test to assess multi-turn persistence. While these simulations provide more dynamic testing environments than traditional benchmarks like AgentBench \cite{liu2025}, they remain limited by the behavioral policy of the simulator. If the simulated user fails to trigger a specific conversational branch, that logic remains unverified. Our framework replaces stochastic ``play-acting'' with an exhaustive Depth-First Search (DFS) traversal of the AgentGraph DSL, providing a deterministic guarantee of topological coverage that simulation cannot reach.

\subsection{Structured Orchestration and Guided Determinism}
Our research builds upon the paradigm of Guided Determinism, where agent reasoning is externalized into design-time state machines or directed graphs, such as LangGraph \yrcite{langgraph} and Agentforce's AgentGraph \yrcite{agentgraph}. Recent studies, such as AGORA \cite{zhang2025} and GEMMAS \cite{lee2025}, have proposed graph-based orchestration for improving agent reproducibility. However, these works primarily focus on execution rather than formal verification. Our methodology is the first to utilize the Graph DSL itself as a ground-truth schema to quantify Structural Adherence, a novel metric measuring the alignment between an agent's runtime transitions and its design-time topology.
\section{Proposed EDGE Framework}

\begin{figure*}[t]
    \centering
    \includegraphics[width=0.9\textwidth]{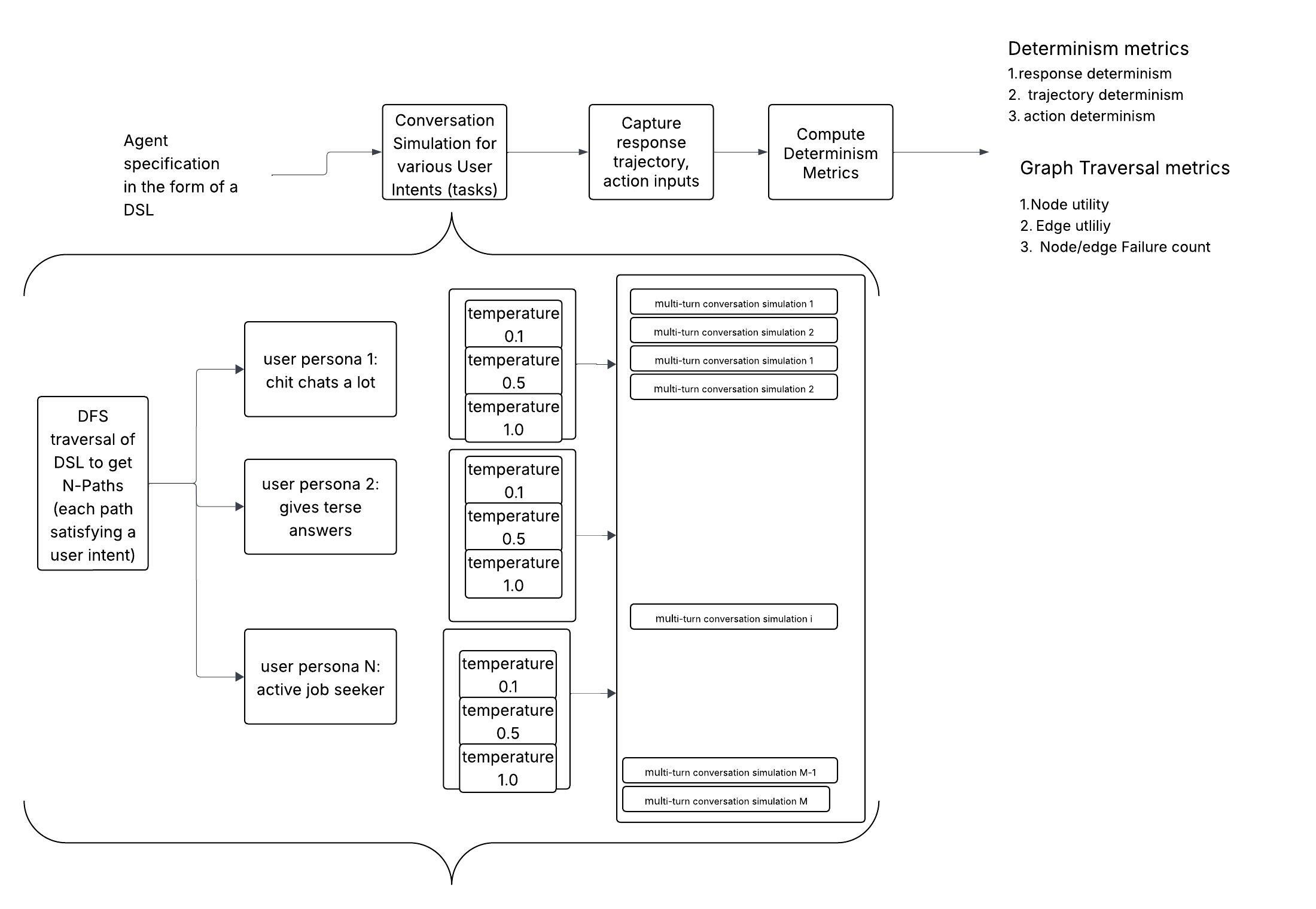}
    \caption{EDGE Framework for Agents}
    \label{fig1}
\end{figure*}

In this section, we describe the various components of the proposed EDGE framework (shown in Figure \ref{fig1}). We specify the components of the agent that needs to be evaluated for determinism using a DSL (domain specific language in json format). The DSL is actually a DAG (Directed Acyclic Graph) where each node in the graph is a sub-agent with a system-2 reasoner capable of managing its own memory, tools and state transitions. An agent is typically designed with the intention of addressing specific tasks for end-users of the agent. For instance, a query from an end-user ``how can I reset my password?'' will translate to a specific path in the AgentGraph  starting from the initial ``Topic or Intent classification'' node to a node that handles ``Login Issues''. Similarly, imagine that each user task that the agent is designed to handle will have its corresponding sequence of nodes to handle. 

Given an agent configuration (in the form of a DSL), the designer persona typically needs to validate whether the agent indeed caters to those tasks in a reliable and consistent manner. Towards that end, we propose the following 3 high-level steps in the EDGE framework for Agents(shown in Figure \ref{fig1}).

\begin{enumerate}
    \item Conversation Simulation for Various user tasks (intents)
    \item Capture agent behavior including response, trajectory and action inputs, etc during simulation
    \item Compute determinism metrics
\end{enumerate}

\subsection{Conversation Simulation}

When a designer specifies an agent through a DSL, the designer has specific intents in mind for the agent to satisfy. The goal of the conversation simulation module is to simulate multi-turn conversations that are about a combination of one or more of these intents. The role of the user is played by an LLM and the agent responds to each turn of the user. In the process, we hope to hit each of the nodes (sub-agents) specified in the DSL that are supposed to work together to satisfy specific intents. To understand this, consider an agent that is designed to handle questions about sports fitness clubs. Let us assume that this agent has the following sub-agents as nodes:

\begin{enumerate}
    \item Topic selector (initial node that routes between subagents)
    \item Find a facility: helps a customer locate a fitness club
    \item Membership and Guest Access: helps with questions about membership plans and guest access
    \item Fitness class schedule: answers questions about class schedules at a particular facility
    \item General Facility information: general information about a known facility but no access to class details
\end{enumerate}

Now, given a query about finding a facility address in New York City and whether one can bring a guest to that facility, we will play the role of a user with that intent and drive the multi-turn conversation using an LLM. The agent in the process of satisfying this user's intent has to atleast hit these three nodes: {Topic Selector, Find a facility, Membership and Guest Access}.

Additionally, we will create multiple conversations with the exact same intent or similar intent by modifying the following aspects:

\begin{itemize}
    \item User persona variation (played by the LLM) e.g chit chatting user, terse user,rude user etc
    \item temperature of the LLM playing the role of the User.
    \item re-writing the intent of the user slightly differently (using an LLM)
\end{itemize}

To be exhaustive about different intents that could be served by the agent, we propose to use Depth First Search (DFS) traversal of the specified agent DSL by the designer which results in N-paths. These N-paths are mapped to M multi-turn conversations serving different intents along with their variations and making the agent traverse through different nodes in the DSL. This dataset becomes the basis for providing metrics on determinism on the agent's behavior.

\subsection{Capture Simulation Details}

In the previous section, we described how we simulate conversations with the agent with different intents where the role of a User is played by an LLM. While doing such simulation, we capture the following details on agent behavior to be able to compute various metrics later on:

\begin{itemize}
    \item User turn (played by an LLM)
    \item Assistant turn ( Agent response for that turn)
    \item Internal states traversed by the agent to respond for that turn
    \item actions invoked along with parameters    
\end{itemize}

With these details captured, we can compute various metrics that provide a view on how confident we can be about the agent's behavior to satisfy different intents.

\subsection{Compute Determinism Metrics}

Based on the M conversations simulated for N-paths in the DSL and captured details for each of those simulated conversations, we present the following two sets of metrics for measuring determinism of the agent.

The first set of determinism metrics are for exact replay of the User's intent and turns. The idea is to see if the task resolution, trajectory of states and actions remain consistent for the exact replay of the user's turns. For instance, we repeat a specific User Intent with the same phrase 3 times and record the agent behavior for those multi-turn conversations. Then, we check for response consistency, node trajectory consistency, and action trajectory consistency between these runs. If the agent behaves similarly for all 3 runs, then these metrics will have high values (shown in Table \ref{tab:determinism_metrics_exact}).

\begin{table*}
    \centering
    \begin{tabular*}{\textwidth}{@{\extracolsep{\fill}}c c p{0.25\textwidth} p{0.45\textwidth}}
        \hline
        Metric & Range & What It Measures & Example \\
        \hline
        Response Consistency & 1--4 &
      Consistency of Agent’s response using LLM-as-a-Judge
 &
If all 3 runs the agent responds with a semantically similar message,
        we get a score of 4. \\

        Node Trajectory Consistency
 & 0-1 & percentage of runs with identical ordered nodes path
&
3 runs: (A→B), (A→B→A), (A→B): 2/3. \\
Node Trajectory Coverage
 & 0-1 &
      Overlap of nodes visited (order-independent)
 &
Sets: {A,B}, {A,B}, {A,B}: 1. \\
Action Trajectory Consistency
 & 0-1 &
      percentage of runs with identical action names (ordered)
 & 3 runs: (KB→CRM),  (KB→CRM→KB→CRM), (KB→CRM)
2/3. \\
Action Trajectory Coverage  & 0-1 &
Overlap of action names called
 &
Sets: {KB,CRM}, {KB,CRM}, {KB,CRM}  1.0
 \\
 Action+Input Trajectory Consistency
 & 0-1 & percentage of runs with identical actions+parameters (ordered)
 & 3 runs:KB(q=X)→CRM(p=A), KB(q=X)→CRM(p=A), KB(q=Y)→CRM(p=B) → 2/3
 \\
 Action+Input Trajectory Coverage & 0-1 &
      Overlap of action+parameter combinations
 &
Sets: {KB(q=X), CRM(p=A)}, {KB(q=X), CRM(p=A)}, {KB(q=Y), CRM(p=B)} → 1/3.\\
        \hline
    \end{tabular*}
    \caption{Determinism Metrics for Exact Replays}
    \label{tab:determinism_metrics_exact}
\end{table*}

The next set of metrics that we propose are for variants for the same task. For instance, instead of asking a particular class schedule for a fitness class at a sports club you might ask a different class schedule at a different club. Since both are in essence the same task we expect the subagents trajectory and action trajectory to be similar but the exact action inputs will be different in these two scenarios. This set of metrics for variants are shown in Table \ref{tab:determinism_metrics_variant}

\begin{table*}
    \centering
    \begin{tabular*}{\textwidth}{@{\extracolsep{\fill}}c c p{0.25\textwidth} p{0.45\textwidth}}
        \hline
        Metric & Range & What It Measures & Example \\
        \hline
        Response Consistency & 1--4 &
      Average score across all variants against each exact-run response

 &
Variants score 4,4,3 → avg 3.67. \\

        Node Trajectory Consistency
 & 0-1 & percentage of variant runs matching base canonical path (ordered)

&
Base: A→B, Variant: A→B → 1.0 \\
Node Trajectory Coverage
 & 0-1 &
      Overlap of nodes across base+variants
 &
Sets: Base:{A,B}, Variant:{A,B} → 1.0\\
Action Trajectory Consistency
 & 0-1 &
      percentage of variant runs calling same action names as base (ordered)

 & 3 runs: (KB→CRM),  (KB→CRM→KB→CRM), (KB→CRM)
2/3. \\
Action Trajectory Coverage  & 0-1 &
Overlap of action names called across base and variant
 &
Sets: {KB,CRM}, {KB,CRM}, {KB,CRM}  1.0
 \\
 Action+Input Trajectory Consistency
 & 0-1 & percentage of runs with identical actions+parameters (ordered)
 & 3 runs:KB(q=X)→CRM(p=A), KB(q=X)→CRM(p=A), KB(q=Y)→CRM(p=B) → 2/3
 \\
 Action+Input Trajectory Coverage & 0-1 &
      Overlap of action+parameter combinations
 &
Sets: {KB(q=X), CRM(p=A)}, {KB(q=X), CRM(p=A)}, {KB(q=Y), CRM(p=B)} → 1/3.\\
        \hline
    \end{tabular*}
    \caption{Determinism Metrics for Variant Replays}
    \label{tab:determinism_metrics_variant}
\end{table*}

\subsection{Metrics computed during Conversation Simulation}

In addition to the determinism metrics that are computed after the conversation simulation, in this section we introduce two metrics that are useful to understand how stable the agent graph is.

\begin{itemize}
    \item Conversation Convergence Rate (CCR): Of the N paths that we wanted to simulate for this agent, this represents the percentage of those that followed the ``intended'' path without any backtracking required during simulation. That is, if we expect to transition through nodes A -> C -> B, we are indeed able to follow that transition. If there is some unexpected transition, we stop and backtrack to the same state from the beginning and try if the intended transition happens again.
    \item Path Completion Rate (PCR): Of the N paths that we wanted to simulate for this agent, this represents the percentage of paths for which we were able to complete the entire set of state transitions (with and without backtracking). There may still be some states that we are not able to reach during simulation, and we want the agent designer to be aware of those states.
\end{itemize}

\subsection{Graph Traversal Metrics}

After the conversation simulation for N-paths using M conversations, we compute the following graph traversal metrics.

\begin{itemize}
    \item Node Utility: How often does a node gets visited during the simulated conversations
    \item Edge Utility: How often does a transition between nodes happen
    \item Average Node Transitions: Mean number of steps per turn in a conversation
    \item Node or Edge failure count: Number of times the agent failed to complete a task at a particular node or edge.
\end{itemize}

\begin{figure}[t]
    \centering
    \includegraphics[width=0.9\columnwidth]{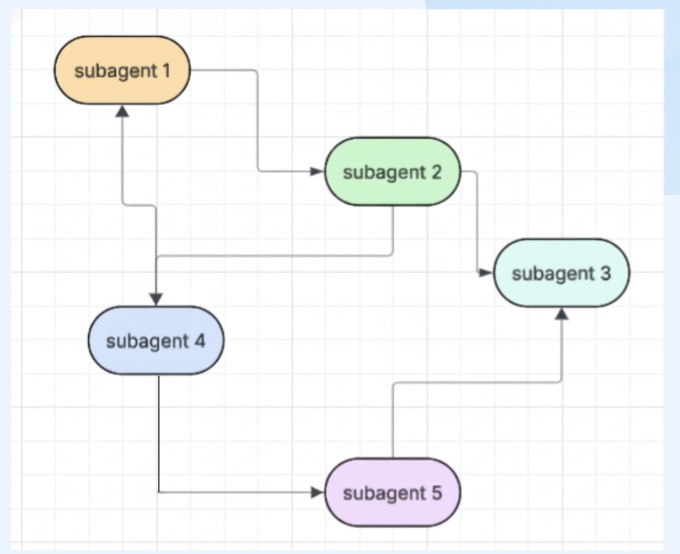}
    \caption{Original intended subagents (nodes) and transitions}
    \label{fig2}
\end{figure}

\begin{figure}[t]
    \centering
    \includegraphics[width=0.9\columnwidth]{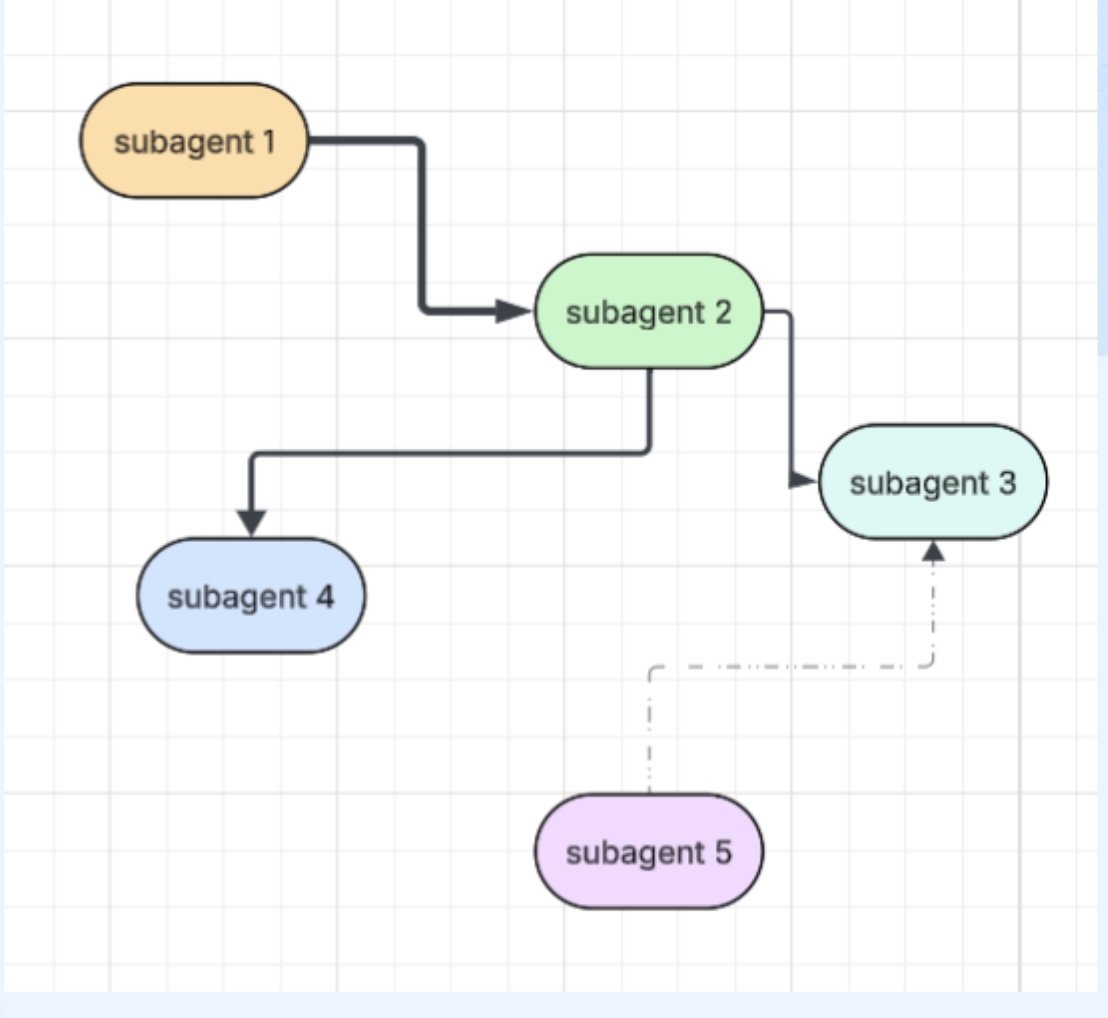}
    \caption{Node and Edge Utility after the conversation simulations}
    \label{fig3}
\end{figure}

The agent designer may have specified a particular graph as shown in Figure \ref{fig2}. However, after the simulations using the computed metrics above we can provide the visualization of actual node utility (subagent visits) and strength of specific edges as shown in Figure \ref{fig3}. This gives an idea of which nodes are important and which ones are least visited for this agent. It can also provide an insight into some missing coverage in the testing of this agent. 

\section{Other Use-Cases}

\subsection{Post-production Evaluation}

In the previous section, we introduced determinism metrics for an agent that a designer of the agent can compute to validate that the agent indeed performs well on expected tasks. Those same metrics could be computed with real-conversations that happened in production after deploying the agent. This time instead of simulating the conversation using an LLM playing the role of a User, we have the real conversations that happened in production and we can replay those for the user turns and record the actual agent response. The same set of metrics proposed in the previous section would help us understand how reliable the agent is in production.

\subsection{Recommending possible end-user turns}

In the previous section through conversation simulation, we saw the node utility and edge utility computed as proportion of times a node or edge is visited. This gives us an idea of what might be the most likely transition given the current state. We could use this information to recommend what the user should most likely say in the next-turn. This is essentially guessing the intent of the user given that we have seen few states up to this point in time.

\section{Experimental Setup}
We used the proposed evaluation framework to evaluate response of three agents:

\begin{itemize}
    \item A car dealer Agent defined in LangGraph framework with 4 nodes
    \item A complex recruitment Agent defined in AgentGraph framework with 6 nodes.
    \item A baseline recruitment Agent defined in AgentGraph with 6 nodes, similar to the above, but with all possible transitions enabled.
\end{itemize}

\subsection{Honda Dealer Agent}

The Honda Dealer agent is a simple 4 node agent tasked to handle primarily 3 intents, namely Scheduling test drives, Warranty look ups, and Answering general car feature related queries based on the Honda Car Manual for ZDX. 

We simulated 30 conversation scenarios by sampling the paths from the agent's graph, including some adversarial scenarios where the user's utterance does not provide the agent with what is needed at a particular turn. Each scenario was replayed 4 times.

\subsection{Recruitment Agent}

The recruitment agent has a complex graph topology, consisting of 6 nodes, each assigned a specific task:

\begin{itemize}
    \item Topic Selector: Root node; responsible for routing the process to the next appropriate node.
    \item Candidate Job Inquiry: Handles candidate's questions about the job.
    \item General Customer Inquiries: Handles candidate's general questions.
    \item Screening Questions: Responsible for walking through a set of screening questions. Transitions to Killer questions in the complex version of the agent happens, only if a candidate passes all screening questions.
    \item Killer Questions: Responsible for walking through a set of questions and recording candidate's answers.
    \item Finalize Interview: Finalizes or terminates the interview, depending on the state of interview, which is updated based on candidate's response to the screening or killer questions.
\end{itemize}

We simulated 136 conversations across all possible DFS paths in this graph, with two variations of the recruitment agent:

Each scenario was replayed 3 times.

\section{Results}
\cref{tab:determinism,tab:node_util} summarize the per-agent determinism and node utilization metrics, respectively, averaged over all simulations.

\begin{table*}
    \centering
    \caption{Determinism metrics.}
    \label{tab:determinism}
    \begin{tabular*}{\textwidth}{@{\extracolsep{\fill}}l c c c}
        \hline
        Metric &   Honda Dealer Agent & Complex Recruitment Agent & Baseline Recruitment Agent \\
        \hline
        Node Trajectory Consistency~(\%)        & 78.89 & 98.89  & 90.19   \\
        \hline
        Node Trajectory Coverage~(\%)   & 76.67 & 98.62  & 91.88   \\
        \hline
        Action Trajectory Consistency~(\%)      & 78.89 & 91.30  & 79.81   \\
        \hline
        Action Trajectory Coverage~(\%) & 66.67 & 93.60  & 89.42   \\
        \hline
        Response Consistency        & 3.13    &  2.92    & 2.25      \\
        \hline
        Average Number of Turns               & 3.2     &  4.96    &  4.96     \\
        \hline
    \end{tabular*}
\end{table*}

\begin{table*}
    \centering
    \caption{Node utilization metrics.}
    \label{tab:node_util}
    \begin{tabular*}{\textwidth}{@{\extracolsep{\fill}}l c c c c}
        \toprule
        Agent & Node\_id & Queries & Visits & Node Util.~(\%) \\
        \midrule
        \multirow{4}{*}{Honda Dealer}
        & greetings & 90 & 137 & 100 \\
        & test\_drive & 46 & 121 & 51.11 \\
        & warranty & 28 & 67 & 31.11 \\
        & ZDX\_faq & 17 & 29 & 18.89 \\
        \midrule
        \multirow{6}{*}{Complex recruitment}
        & killer\_questions & 540 & 2172 & 100\\ 
        & candidate\_job\_inquiry & 400	& 1625 & 74.07\\
        & topic\_selector & 376	& 1988 & 69.63\\
        & general\_customer\_inquiries & 300 &	1580 &	55.56\\
        & screening\_questions & 4 & 9 &	0.74\\
        & off\_topic & 3 &	7 &	0.56 \\
        \midrule
        \multirow{6}{*}{Baseline recruitment}
        & killer\_questions & 539 &	2169 &	99.81 \\
        & topic\_selector & 376 &	1988 &	69.63 \\
        & candidate\_job\_inquiry & 407	& 1642 &	75.37\\
        & general\_customer\_inquiries & 300 &	1580 &	55.56\\
        & screening\_questions & 110 &	377 &	20.37\\
        & finalize\_interview & 13 &	22 & 2.41  \\
        \bottomrule
    \end{tabular*}
\end{table*}

\begin{figure}[ht]
\vskip 0.2in
\begin{center}
\centerline{\includegraphics[width=\columnwidth]{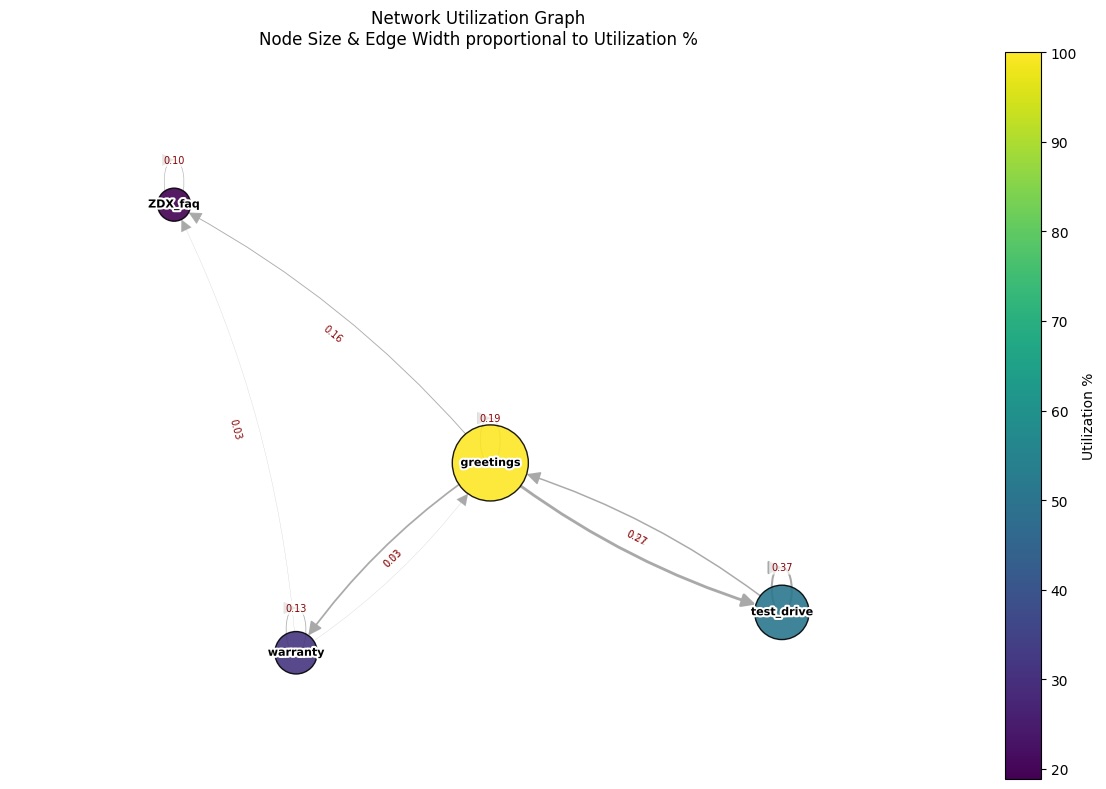}}
\caption{Honda Dealer Agent defined in LangGraph framework.}
\label{honda-agent-graph}
\end{center}
\vskip -0.2in
\end{figure}

\begin{figure}[ht]
\vskip 0.2in
\begin{center}
\centerline{\includegraphics[width=\columnwidth]{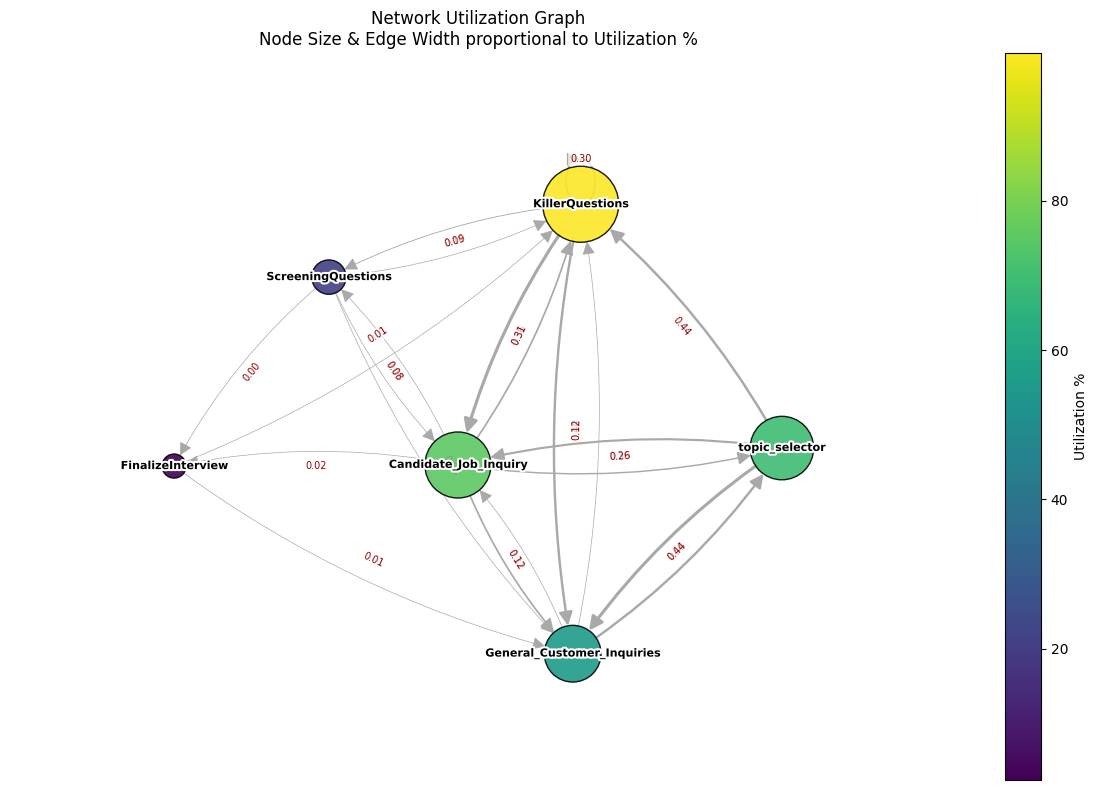}}
\caption{Baseline Recruitment Agent defined in AgentGraph framework.}
\label{recruitment-agent-graph}
\end{center}
\vskip -0.2in
\end{figure}

\cref{honda-agent-graph,recruitment-agent-graph} show the connectivity pattern among nodes for the two agents, with the relative sizes of the nodes and the widths of the edges are proportional to their average utilization.

\section{Discussion}
According to the determinism metrics presented in Table~\ref{tab:determinism}, the Honda Dealer agent exhibits lower determinism in both the order and overlap of visited nodes and, consequently, in the actions selected by those nodes, compared to both variants of the Recruitment agent. The superior performance of the Recruitment agent can be attributed to the presence of a dedicated topic selector node, which consistently routes the process after each turn of conversation. Despite high Trajectory Consistency, the Recruitment agent falls short in generating consistent responses, as indicated from the relatively low values for Response consistency, in both variants of the agent. This observation highlights that adhering to the expected path and selecting actions consistently do not necessarily guarantee semantic consistency in the agent’s response. 

The complex variant of the Recruitment agent outperforms its baseline variant, likely because its graph topology is less dense. The sparse topology is consistent with the greater differences in the node utilization percentages observed in the complex agent compared to the baseline. By removing certain edges, we constrain the agent’s decision space, thus balancing reactiveness and proactiveness and improving determinism.

\section{Conclusion}

In this work, we presented EDGE, a framework for systematically evaluating determinism and behavioral consistency in graph-based agentic systems through conversation simulation and replay. Using the agent’s graph configuration as a ground-truth specification, EDGE enables reproducible testing across all intended conversational paths and provides interpretable metrics for analyzing response , node, and action trajectory consistency.

Our experimental results highlight the value of structured graph design, particularly in the Recruitment agent. The complex Recruitment agent shows higher determinism and more stable execution than the baseline version. This comes from using more deliberate and constrained transitions rather than a broadly permissive graph.

The Honda dealer agent implemented using LangGraph serves primarily as an illustrative example that shows that EDGE is not limited to a single authoring framework. This showcases the generality of the approach. Overall, EDGE offers a practical way to evaluate complex agent workflows and see how consistently they behave in practice.

\section*{Impact Statement}

This paper presents the work whose goal is to advance the evaluation of agentic systems in Machine Learning. The proposed framework supports improved assessment of determinism and enables systematic simulation of conversations in graph-based agent workflows. We do not identify any broader societal impacts that require specific discussion.

\nocite{langley00}

\bibliography{example_paper}

@misc{khatchadourian2025,
      title={LLM Output Drift: Cross-Provider Validation \& Mitigation for Financial Workflows}, 
      author={Raffi Khatchadourian and Rolando Franco},
      year={2025},
      eprint={2511.07585},
      archivePrefix={arXiv},
      primaryClass={cs.LG},
      url={https://arxiv.org/abs/2511.07585}, 
}

@misc{liu2025,
      title={AgentBench: Evaluating LLMs as Agents}, 
      author={Xiao Liu and Hao Yu and Hanchen Zhang and Yifan Xu and Xuanyu Lei and Hanyu Lai and Yu Gu and Hangliang Ding and Kaiwen Men and Kejuan Yang and Shudan Zhang and Xiang Deng and Aohan Zeng and Zhengxiao Du and Chenhui Zhang and Sheng Shen and Tianjun Zhang and Yu Su and Huan Sun and Minlie Huang and Yuxiao Dong and Jie Tang},
      year={2025},
      eprint={2308.03688},
      archivePrefix={arXiv},
      primaryClass={cs.AI},
      url={https://arxiv.org/abs/2308.03688}, 
}

@online{agentgraph,
  author       = {Mui, Phil},
  title        = {Agentforce’s Agent Graph: Toward Guided Determinism with Hybrid Reasoning},
  year         = {2025},
  month        = oct,
  day          = {20},
  url          = {https://engineering.salesforce.com},
  note         = {Salesforce Engineering Blog}
}

@inproceedings{zhang2025,
    title = "Unifying Language Agent Algorithms with Graph-based Orchestration Engine for Reproducible Agent Research",
    author = "Zhang, Qianqian  and
      Liao, Jiajia  and
      Ying, Heting  and
      Ma, Yibo  and
      Shen, Haozhan  and
      Li, Jingcheng  and
      Liu, Peng  and
      Zhang, Lu  and
      Fang, Chunxin  and
      Lee, Kyusong  and
      Xu, Ruochen  and
      Zhao, Tiancheng",
    editor = "Mishra, Pushkar  and
      Muresan, Smaranda  and
      Yu, Tao",
    booktitle = "Proceedings of the 63rd Annual Meeting of the Association for Computational Linguistics (Volume 3: System Demonstrations)",
    month = jul,
    year = "2025",
    address = "Vienna, Austria",
    publisher = "Association for Computational Linguistics",
    url = "https://aclanthology.org/2025.acl-demo.11/",
    doi = "10.18653/v1/2025.acl-demo.11",
    pages = "107--117",
    ISBN = "979-8-89176-253-4"
}

@inproceedings{lee2025,
    title = "{GEMMAS}: Graph-based Evaluation Metrics for Multi Agent Systems",
    author = "Lee, Jisoo  and
      Chang, Raeyoung  and
      Kwon, Dongwook  and
      Singh, Harmanpreet  and
      Verma, Nikhil",
    editor = "Potdar, Saloni  and
      Rojas-Barahona, Lina  and
      Montella, Sebastien",
    booktitle = "Proceedings of the 2025 Conference on Empirical Methods in Natural Language Processing: Industry Track",
    month = nov,
    year = "2025",
    address = "Suzhou (China)",
    publisher = "Association for Computational Linguistics",
    url = "https://aclanthology.org/2025.emnlp-industry.106/",
    doi = "10.18653/v1/2025.emnlp-industry.106",
    pages = "1522--1532",
    ISBN = "979-8-89176-333-3"
}

@misc{openevals,
  author = {LangChain, Inc},
  title = {OpenEvals: Evaluating LLM Applications},
  year = {2025},
  url = {https://github.com/langchain-ai/openevals}
}

@misc{agentevals,
  author = {LangChain, Inc},
  title = {AgentEvals: Evaluating Agent Trajectories},
  year = {2025},
  url = {https://github.com/langchain-ai/agentevals}
}

@misc{langsmith,
  author       = {LangChain, Inc},
  title        = {LangSmith: Platform for Building and Monitoring Production-Grade LLM Applications},
  year         = {2025},
  url          = {https://docs.langchain.com/langsmith/home},
}

@misc{langgraph,
  author       = {LangChain, Inc},
  title        = {LangGraph: Build LLM Agents as Graph},
  year         = {2025},
  url          = {https://docs.langchain.com/oss/python/langgraph/overview},
}

@misc{arizephoenix,
  author       = {Arize-AI},
  title        = {Phoenix: AI Observability and Evaluation},
  year         = {2025},
  url          = {https://arize.com/docs/phoenix},
}

@misc{yehudai2025,
      title={Survey on Evaluation of LLM-based Agents}, 
      author={Asaf Yehudai and Lilach Eden and Alan Li and Guy Uziel and Yilun Zhao and Roy Bar-Haim and Arman Cohan and Michal Shmueli-Scheuer},
      year={2025},
      eprint={2503.16416},
      archivePrefix={arXiv},
      primaryClass={cs.AI},
      url={https://arxiv.org/abs/2503.16416}, 
}

@inproceedings{langley00,
 author    = {P. Langley},
 title     = {Crafting Papers on Machine Learning},
 year      = {2000},
 pages     = {1207--1216},
 editor    = {Pat Langley},
 booktitle     = {Proceedings of the 17th International Conference
              on Machine Learning (ICML 2000)},
 address   = {Stanford, CA},
 publisher = {Morgan Kaufmann}
}
\bibliographystyle{icml2025}

\newpage
\appendix
\onecolumn

\end{document}